\documentclass{article}

\usepackage[final]{neurips_2024}

\usepackage{graphicx} 

\usepackage[utf8]{inputenc} 
\usepackage[T1]{fontenc}    

\usepackage{hyperref}       
\usepackage{url}            

\usepackage{xcolor}         

\usepackage{graphicx}       
\usepackage{caption}        
\usepackage{subfig}
\usepackage{pdflscape}      
\usepackage{placeins}       
\usepackage{float}          
\usepackage{circledsteps} 
\usepackage{hyperref}
\usepackage{tikz}
\usepackage{amsmath,amssymb,amsfonts}
\usepackage{lipsum}
\usepackage{adjustbox}
\usepackage{graphicx}
\usepackage{placeins}
\usepackage{lipsum}
\usepackage{soul,color}
\usepackage{pdflscape}
\usepackage{rotating}
\usepackage{makecell}
\usepackage{placeins}
\usepackage{float}
\usepackage{booktabs}
\usepackage{caption}
\usepackage{subcaption}

\usepackage{microtype}
\usepackage{amsmath} 
\usepackage{multirow}
\usepackage{pifont}
\usepackage{xcolor}

\DeclareMathAlphabet{\pazocal}{OMS}{zplm}{m}{n}

\newcommand{\Lb}{\pazocal{L}}
\DeclareMathOperator{\vect}{vec}

\DeclareMathOperator{\Tr}{Tr}

\newcommand{\best}[1]{\textbf{#1*}}
\newcommand{\cmark}{\ding{51}}%
\newcommand{\xmark}{\ding{55}}%

\usetikzlibrary{arrows.meta}
\tikzset{%
  >={Latex[width=1.4mm,length=2.5mm]},
            base/.style = {rectangle, rounded corners, draw=black,minimum width=2cm, minimum height=1cm,text centered},
  rect/.style = {base, fill=orange!20},
  Start/.style = {base, fill=green!10},
  basemodel/.style = {base, fill=blue!15},
  loss/.style = {base, fill=red!15},
}
\usepackage{xparse}
\usepackage{tikz}           
\usepackage{tikz-cd}        
\usepackage{tikz-3dplot}    
\usetikzlibrary{decorations.pathreplacing, shapes.geometric, arrows.meta, positioning, calc}
\usetikzlibrary{decorations.markings}
\usetikzlibrary{tikzmark,arrows.meta,bending,calc}

\usepackage{pgfplots}       
\usepackage{pgfplotstable}  
\pgfplotsset{compat=1.17}   

\usepackage{booktabs}       
\usepackage{makecell}       
\usepackage{multirow}       
\usepackage{threeparttable} 
\usepackage{adjustbox}      

\usepackage{boldline}

\usepackage{amsmath, amsfonts, amssymb, mathrsfs} 
\usepackage{amsthm}

\usepackage{bm}                        
\usepackage{euscript}                  
\usepackage{nicefrac}                  
\usepackage{cancel}                    

\usepackage{microtype}      
\usepackage{soul}           
\usepackage{xspace}         
\usepackage{fancyvrb}       

\usepackage{enumitem}       
\usepackage{extpfeil}       
\usepackage{rotating}       
\usepackage{pifont}         
\usetikzlibrary{shadows,arrows.meta,positioning,decorations.markings, decorations.pathmorphing}

\tikzstyle{block} = [rectangle, rounded corners, minimum width=3cm, minimum height=1cm,text centered, draw=black, fill=blue!10]
\tikzstyle{arrow} = [thick, ->, >=stealth]

\makeatletter
\pgfdeclarearrow{
  name=ChainLink,
  parameters={},
  setup code={
    \pgfarrowssettipend{0pt}
    \pgfarrowssetbackend{-8pt}
    \pgfarrowssetlineend{-5pt} 
  },
  drawing code={
    \pgfsetlinewidth{1.5pt}
    \pgfsetroundcap
    \pgfsetroundjoin
    \pgfpathmoveto{\pgfpoint{0pt}{-1.5pt}}
    \pgfpathlineto{\pgfpoint{-5pt}{-1.5pt}}
    \pgfpathlineto{\pgfpoint{-5pt}{1.5pt}}
    \pgfpathlineto{\pgfpoint{0pt}{1.5pt}}
    \pgfpathclose
    \pgfusepathqstroke
    \pgfsetlinewidth{0.8pt}
    \pgfsetcolor{white}
    \pgfpathmoveto{\pgfpoint{-2.3pt}{-2pt}}
    \pgfpathlineto{\pgfpoint{-2.7pt}{2pt}}
    \pgfusepathqstroke
  },
  defaults={}
}
\makeatother

\title{Scalable Dynamic Mixture Model with Full Covariance for Probabilistic Traffic Forecasting}

\author{
Seongjin Choi\\
University of Minnesota\\
Minneapolis, MN, USA\\
\texttt{chois@umn.edu}\\
\And
Nicolas Saunier\\
Polytechnique Montreal\\
Montreal, QC, Canada\\
\texttt{nicolas.saunier@polymtl.ca}\\
\And
Vincent Zhihao Zheng\\
McGill University\\
Montreal, QC, Canada\\
\texttt{zhihao.zheng@mail.mcgill.ca}\\
\And
Martin Trépanier\\
Polytechnique Montreal\\
Montreal, QC, Canada\\
\texttt{mtrepanier@polymtl.ca}\\
\And
Lijun Sun\\
McGill University\\
Montreal, QC, Canada\\
\texttt{lijun.sun@mcgill.ca}
}

\begin{document}

\maketitle

\begin{abstract}
Deep learning-based multivariate and multistep-ahead traffic forecasting models are typically trained with the mean squared error (MSE) or mean absolute error (MAE) as the loss function in a sequence-to-sequence setting, simply assuming that the errors follow an independent and isotropic Gaussian or Laplacian distributions. However, such assumptions are often unrealistic for real-world traffic forecasting tasks, where the probabilistic distribution of spatiotemporal forecasting is very complex with strong concurrent correlations across both sensors and forecasting horizons in a time-varying manner. In this paper, we model the time-varying distribution for the matrix-variate error process as a dynamic mixture of zero-mean Gaussian distributions. To achieve efficiency, flexibility, and scalability, we parameterize each mixture component using a matrix normal distribution and allow the mixture weight to change and be predictable over time. The proposed method can be seamlessly integrated into existing deep-learning frameworks with only a few additional parameters to be learned. We evaluate the performance of the proposed method on a traffic speed forecasting task and find that our method not only improves model performance but also provides interpretable spatiotemporal correlation structures. 
\end{abstract}

\section{Introduction}
The unprecedented availability of traffic data and advances in data-driven algorithms have sparked considerable interest and rapid developments in traffic forecasting. State-of-the-art deep learning models such as DCRNN \citep{li2017diffusion}, STGCN \citep{yu2017spatio}, Graph-Wavenet \citep{wu2019graph}, and Traffic Transformer \citep{cai2020traffic}, have demonstrated superior performance for traffic speed forecasting over classical methods. Deep neural networks have shown clear advantages in capturing complex non-linear relationships in traffic forecasting; for example, in the aforementioned models, Graph Neural Networks (GNNs), Recurrent Neural Networks (RNNs), Gated Convolutional Neural Networks (Gated-CNNs), and Transformers are employed to capture those unique spatiotemporal patterns in traffic data \citep{li2015trend}.

When training deep learning models, conventional loss functions include Mean Squared Error (MSE) and Mean Absolute Error (MAE). This convention of using MSE or MAE is based on the assumption that the errors follow either an independent Gaussian distribution or a Laplacian distribution. If the errors are assumed to follow an independent and isotropic zero-mean Gaussian distribution, the Maximum Likelihood Estimation (MLE) corresponds to minimizing the MSE. Likewise, the MLE corresponds to minimizing MAE if the errors are assumed to follow an isotropic zero-mean Laplacian distribution. However, the assumption of having uncorrelated and independent errors does not hold in many real-world applications. 

Particularly, both spatial and temporal correlation is prevalent in traffic data. It is likely that traffic conditions in one location may have an impact on traffic conditions in nearby locations, and the error for a given time period may be correlated with the error for previous or future time periods. Ignoring these spatial and temporal dependencies can lead to a biased forecasting model. Additionally, real-world data is often highly nonlinear and nonstationary, which can make it difficult to fit a single multivariate Gaussian distribution to the high-dimensional data. This is particularly true for traffic data, which can be highly dynamic and exhibit substantial and abrupt changes over time, such as during the transition from free flow to traffic congestion. 

\begin{figure}[!t]
    \centering
    \includegraphics[width=0.6\textwidth]{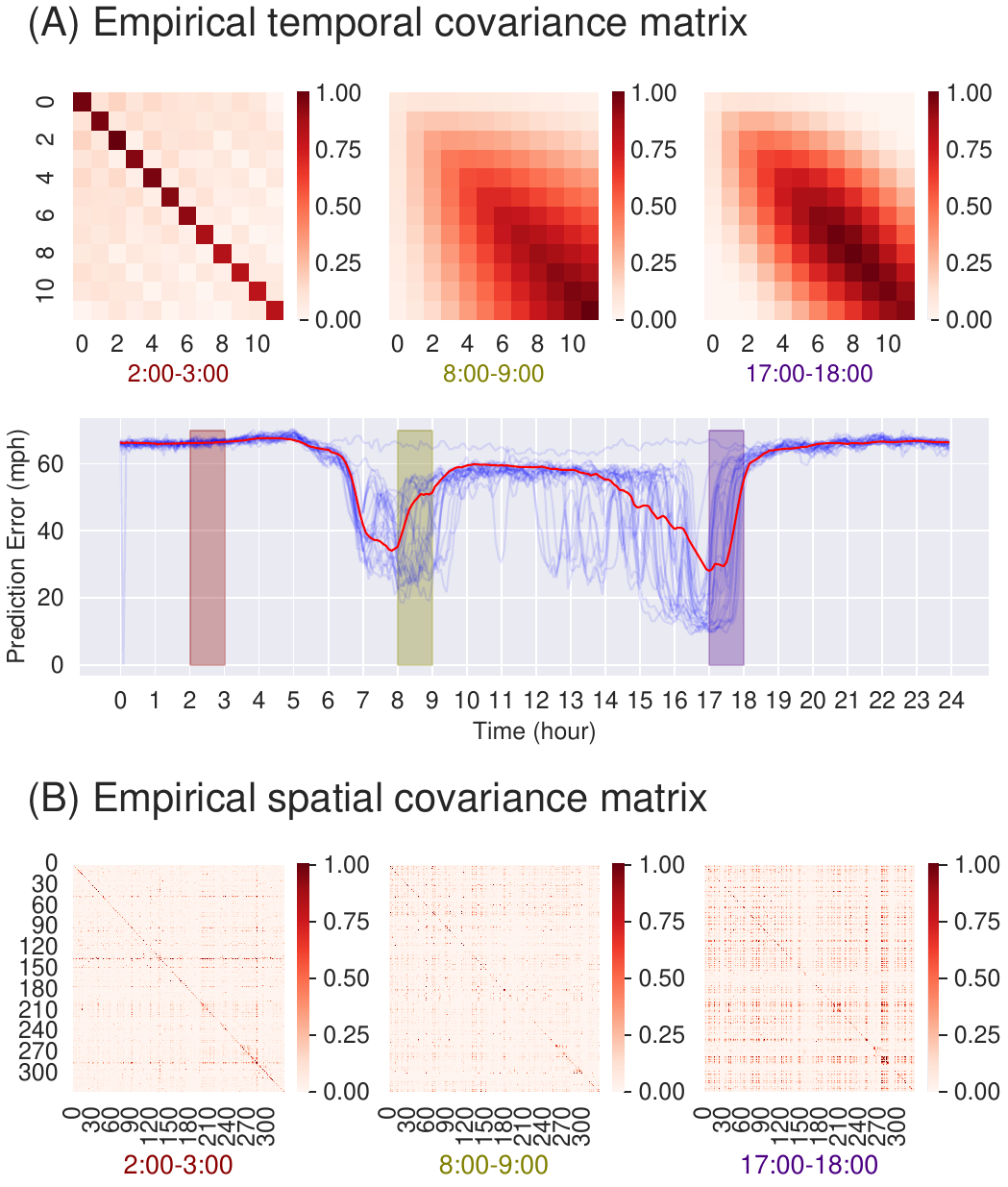}
    \caption{
    Empirical results based on forecasting results from Graph Wavenet \cite{wu2019graph} with PEMS-BAY data
    \textbf{(A)} temporal (i.e., over different prediction horizons) covariance matrices of \textit{Sensor \# 6} at different time-of-days (2:00-3:00, 8:00-9:00, and 17:00-18:00). 
    \textbf{(B)} spatial (i.e., over different sensors) covariance matrices of \textit{12-step-ahead} prediction at different time-of-days (2:00-3:00, 8:00-9:00, and 17:00-18:00). 
    }
    \label{fig:fig1}
\end{figure}

The straightforward idea to resolve the limitations related to using MSE and MAE is to properly model the underlying distribution of the forecasting error. The challenges are that the forecasting error is spatiotemporally correlated (i.e., non-diagonal entries should have values), and the distribution of forecasting error is often time-varying and multimodal. To demonstrate these challenges, we show the empirical covariance matrices based on the forecasting result from Graph Wavenet \cite{wu2019graph} using the PEMS-BAY dataset (325 sensors and 12-step prediction with 5-minute intervals). Specifically, in Figure \ref{fig:fig1} (A), we demonstrate the empirical temporal covariance matrix of one sensor (the 6th sensor among 325 sensors) to focus solely on the temporal correlation. Likewise, in Figure \ref{fig:fig1} (B), we demonstrate the empirical spatial covariance matrix of one prediction time step (12-step-ahead prediction) to focus solely on the spatial correlation. The empirical results show that the covariance matrices at different times of the day show different and distinct correlation structures, meaning that the distribution is time-varying and multimodal, with a strong correlation in both spatial and temporal dimensions.



In this study, we propose to model the forecasting errors with time-varying distribution to consider the spatiotemporal correlation, and train the model by maximizing the likelihood function in addition to conventional loss function. Specifically, we use a dynamic (i.e., time-varying) mixture density network as a flexible structure by allowing the full covariance matrix to capture the spatiotemporal dependencies.
The fundamental challenge in this approach is the high computational cost due to the large $NQ\times NQ$ spatiotemporal covariance matrix for the forecasting error, where $N$ is the number of sensors and $Q$ is the forecasting horizon. 
To address this, we reparameterize the covariance in each mixture component as a Kronecker product of two separate spatial and temporal covariance matrices.
In order to efficiently learn the covariance matrix in the deep learning model, we parameterize the spatial and temporal covariance matrices using the Cholesky factorization of the precision matrices (inverse of the covariance matrix). 
Overall, the proposed method can be seamlessly integrated into existing deep-learning frameworks with only a few additional parameters to be learned. 

{
The key contributions of this paper are as follows:
\begin{itemize}[
  align=left,
]
    \item We introduce a novel framework to model forecasting errors as a dynamic mixture of matrix-variate Gaussian distributions, effectively capturing time-varying, multimodal, and spatiotemporal dependencies. 
    \item We propose a scalable parameterization for the spatiotemporal covariance matrix and derive an analytical expression for the proposed loss function to ensure computational efficiency.
    \item The proposed method integrates seamlessly into existing deep learning models with minimal additional parameters. 
    \item The model demonstrates improved forecasting performance while providing interpretable spatiotemporal correlation structures, enhancing insights into traffic dynamics. 
\end{itemize}
}

\section{Related Works}

%

There are mainly two approaches for modeling time-varying probabilistic distribution in statistics and machine learning.
The first approach aims to model time-varying covariance matrix to capture dynamicity and spatiotemporal correlation. One method is the Wishart process \citep{wilson2010generalised}, which is a stochastic process that models the time-varying covariance matrix of a multivariate time series. 
Another widely-used method is the GARCH (Generalized Autoregressive Conditional Heteroskedasticity) model \citep{bauwens2006multivariate}, which have been widely used to model the time-varying variances of time series data, especially widely used for the finance data. GARCH models assume that the variance of the errors is a function of past errors, allowing for the capture of dynamic volatility patterns. 
%
Moreover, recent studies combining statistical approaches with deep learning also align in this category. \citet{salinas2020deepar} proposed DeepAR to produce probabilistic forecasting for time-series. The given neural network outputs the parameters of Gaussian distribution (the mean and the variance), and multiple time series could be trained jointly with shared covariates. \citet{salinas2019high} employed a Gaussian copula with a low-rank structure for time-varying covariance learning with neural networks.
However, these models assume that the distribution of the errors follows a unimodal distribution, while the distribution may be multimodal for spatiotemporally correlated errors.

Another approach is to use a mixture of distributions focusing on multimodality. 
{
Examples in this approach typically use a mixture of distributions, also known as mixture models, to capture the multimodality in the data. Theoretical support for using mixture models in traffic data can be found in the literature where they are often employed to model heterogeneous data that contains multiple underlying populations or behaviors. For instance, studies in traffic research have shown that mixture models provide a more flexible framework for capturing the variations and irregularities in traffic flows and speeds (refer to \cite{vaughan1970distribution,kerner1997experimental,couton1997application,park2010bayesian}). These models allow for the representation of different traffic regimes as separate components within the mixture, each with its own statistical properties. There are several previous studies in time-series analysis, including traffic data analysis, that aim to capture multimodality and heterogeneity in datasets. 
}
One example is \citet{wong2000mixture} which proposed the Mixture Autoregressive (MAR) model to model non-linear time series by using a mixture of Gaussian autoregressive components. Also, a series of works use the Mixture Density Network (MDN) \cite{bishop1994mixture} by combining the mixture idea with neural networks. \citet{nikolaev2013time} presented the recurrent mixture density network combined with GARCH (RMDN-GARCH) for time-varying conditional density estimation. \citet{ellefsen2019mixture} also analyzed the prediction generated by mixture density recurrent neural networks (MD-RNN) and tested the ability to capture multimodality by using the mixture density network. However, these studies usually use MDN to model univariate time series prediction. 
%
{
There are a few other examples such as \cite{mao2023gmdnet,chen2021short,li2023xrmdn} that use multivariate Gaussian distribution as the component distribution of MDN, however, these studies typically assume diagonal covariance matrix which cannot capture the spatiotemporally correlated structure in the error distributions. Other studies \citep{liu2023gaussian,cai2020spatiotemporal,bayati2020gaussian} model the distribution as a multivariate Gaussian distribution with full covariance using the Gaussian process, however, this approach requires a pre-defined kernel structure which can be a lot more restrictive.
}

Our idea is to leverage the advantages of both approaches while keeping the model scalable to train without heavy computation. We use a dynamic mixture of zero-mean Gaussian distribution to characterize the distribution of forecasting errors, where the mixture weight is time-varying, while the covariance matrices in component distributions are fixed. Also, we propose a decomposition method using the Kronecker product to keep the scalability without further assuming the characteristics (such as low-rank-ness as in \citet{salinas2019high}) of the covariance matrix.


\section{Methodology}
\subsection{Background}
First, we briefly give out the formulation of the problem, i.e., short-term traffic flow or speed forecasting. Given a time window $t$, we denote by $v_t^n$ the average observations (i.e., speed or flow) of the $n$-th sensor. The observations in the whole network at the given time window are denoted by vector $\mathbf{x}_t = \left[v_t^1,v_t^2,\cdots,v_t^N\right]^{\top}\in\mathbb{R}^N$. 
Given the $P$ historical observations ${X}_t = \left[\mathbf{x}_{t-P+1}, \cdots , \mathbf{x}_{t} \right]\in\mathbb{R}^{N \times P}$, we aim to predict the speed of $N$ sensors for  the next $Q$ steps, i.e., ${Y}_t \in \mathbb{R}^{N \times Q}$ as a spatiotemporal speed matrix. 

Conventional training of deep-learning models for short-term traffic speed forecasting problems can be regarded as learning of a mean-prediction function $f_{M}$ by minimizing MSE (the loss $\mathcal{L}_{\text{MSE}}$):
\begin{equation}
    \begin{split}
    & f^*_{M} = \arg \min_{f_{M}} \mathcal{L}_{\text{MSE}},  \\
    & \mathcal{L}_{\text{MSE}} = \|Y_t - M_t\|^2_F, \\
    \end{split}
\end{equation}
where $M_t=f_{M}(X_t) \in \mathbb{R}^{N \times Q}$ is a matrix of the expected future traffic speed.
The convention of using MSE loss for training deep learning models for traffic speed forecasting assumes that the error, $R_t=Y_t-M_t \in \mathbb{R}^{N \times Q}$, follows a zero-mean Gaussian distribution with isotropic noise:
\begin{equation}
    \begin{split}
    \operatorname{vec}({R}_t) \sim \mathcal{N}(0, {\Sigma = \sigma^2 I_{NQ}}),
    \end{split}
    \label{eq:res_conv}
\end{equation}

{
Using MSE as a loss function for a training deep-learning model corresponds to maximizing the likelihood (using negative log-likelihood as loss):
\begin{equation}
\begin{split}
    - \log p(Y_t | X_t) 
    & =  - \log \left( \frac{1}{\sqrt{2\pi |\Sigma|}} \exp{\left( - \frac{1}{2}  \vect(R_t)^\top \Sigma^{-1} \vect(R_t) \right) } \right) \\
    & =  \frac{1}{2}\log (2\pi)  + \frac{1}{2} \log |\Sigma| + \frac{1}{2} \vect(R_t)^\top \Sigma^{-1} \vect(R_t) \\ 
    & =  \frac{1}{2}\log (2\pi)  + \frac{NQ}{2} \log \sigma^2 + \frac{1}{2 \sigma^2} \vect(R_t)^\top \vect(R_t) \\ 
    & \propto \vect(R_t)^2 = \|Y_t - M_t\|^2_F  = \mathcal{L}_{\text{MSE}}, \\    
\end{split}
\end{equation}
since the terms other than $\vect(R_t)^2 $ are constant.
}

\subsection{Learning Spatiotemporal Error Distribution with Dynamic Mixture}
%

In this study, we aim to model the distribution of forecasting error, and train a given deep-learning-based traffic forecasting model with a redesigned loss function. We use a dynamic mixture of zero-mean Gaussian distribution, where only the mixture weights are time-varying.
%
Also, we explicitly model the distribution of error with spatiotemporal correlation and directly train the model by explicitly using the negative log-likelihood loss as follows:
\begin{equation}\label{eq:loss}
  \begin{split}
  &\mathcal{L}_{\text{DynMix}}  = (1-\rho) \mathcal{L}_{\text{MSE}} + \rho \mathcal{L}_{\text{NLL}},\\
  \end{split}
\end{equation}
where 
$\mathcal{L}_{\text{NLL}}$ is the negative log-likelihood of the forecasting errors and $\rho$ is the weight parameter. 
{
Essentially, different loss functions suggest different ways to model the conditional probability of the prediction error between future traffic states ($Y_t$) and prediction ($M_t$) given the historical observation ($X_t$). As a result, using MSE alone as a loss function and using the proposed dynamic mixture of zero-mean Gaussian distribution are founded on the same theoretical background.
}

One intuitive and interpretable solution to model a complex multivariate distribution is to use a \textit{mixture of Multivariate Gaussian distributions}, which can represent or approximate practically any distribution of interest while being intuitive and easy to handle \citep{bishop1994mixture,hara2018network}. Furthermore, we use a \textit{dynamic mixture of multivariate zero-mean Gaussian distribution with a full covariance matrix} to model the distribution of spatiotemporally correlated errors and assume that the probability density function of the error can be formulated as a mixture of $K$ multivariate zero-mean Gaussian with dynamic (i.e., time-varying) mixture weights $\omega^k_t$:
\begin{equation}
 \operatorname{vec}\left({R}_t\right)  \sim \sum_{k=1}^K \omega_t^k \cdot \mathcal{N} \left({0},\Sigma^k\right), 
\end{equation}
where ${\Sigma^k} \in \mathbb{R}^{NQ \times NQ}$ is the time-independent spatiotemporal covariance matrix for $k$-th component, and $\omega_t^k\ge 0$ is the time-varying weight (proportion) for the $k$-th covariance matrix, which should be sum to 1, i.e., $\sum_{k=1}^K \omega_t^k = 1$, for any $t$. Here, note that the mixture weight, $\omega_t^k$, is time-dependent (or input-dependent; $\omega_t^k = f_{\omega}(X_t)$), while the covariance matrix in the component distribution $\Sigma^k$ is not. This representation allows modeling the time-varying distribution of error distribution while remaining intuitive and interpretable by analyzing the component distribution. 
{
If the model fully captures the non-stationary residual process, the mixture weights converge to a single dominant mode, and the covariance matrices become diagonal with small variances.
}

However, this approach has a critical issue in computing the log-likelihood of this distribution, which requires high computation cost due to the large size of ${\Sigma^k}$ (i.e., $NQ\times NQ$). Directly learning such a large spatiotemporal covariance matrix (${\Sigma^k} \in \mathbb{R}^{NQ \times NQ}$) is infeasible when $N$ or $Q$ becomes large, since computing the negative log-likelihood involves calculating the inverse and the determinant of the covariance matrix. For example, the widely-used PEMS-BAY traffic speed data consists of observations from 325 sensors, and the forecasting horizon is usually set to 1 hour (i.e., 12 timesteps with a 5-minute interval). This would result in $N=325$ and $Q=12$, and the size of $\Sigma^k$ becomes ${3900 \times 3900}$. 

To address the scalability issue, we propose a decomposition method that parameterizes each $\Sigma^k$ as a Kronecker product of spatial covariance matrix $\Sigma_N^k$ and temporal covariance matrix $\Sigma_Q^k$:
\begin{equation}
R_t \sim \sum_{k=1}^K \omega_t^k \cdot \mathcal{N} \left({0},\Sigma^k=\Sigma^k_Q \otimes \Sigma^k_N\right),
\end{equation}
where $\Sigma^k_N \in \mathbb{R}^{N \times N}$ and $\Sigma^k_Q \in \mathbb{R}^{Q \times Q}$ are spatial and temporal covariance for the $k$-th component. 

{
This approach assumes that the spatial and temporal covariances are \textit{separable}, but not necessarily independent. 
{
There has been empirical evidence from previous studies that covariance matrices of traffic data, or spatiotemporal data, can be assumed to follow a Kronecker structure \citep{lei2022bayesian, greenewald2015robust}.
}
Training the full covariance matrix ($\Sigma$) without such decomposition might be an alternative to capture the \textit{spatiotemporally} correlated errors, however, this brings another problem of huge computational complexity as discussed earlier, since computing the negative log-likelihood involves the inverse and determinant operations of the covariance matrix, which usually is $O(n^3)$ in time complexity. 
}

This representation is equivalent to assuming the error $R_t$ follows a dynamic mixture of zero-mean matrix normal distribution:
\begin{equation}
R_t \sim \sum_{k=1}^K \omega_t^k \cdot \mathcal{MN} ({0},\Sigma^k_N,\Sigma^k_Q),
\end{equation}
where $\mathcal{MN}$ represents a zero-mean matrix normal distribution with probability density function:
\begin{equation}
\begin{split}
& p^k_{MN}(X|{0},\Sigma^k_Q,\Sigma^k_N) =  \frac{\exp{\left(
- \frac{1}{2} \Tr \left[ {(\Sigma^k_Q)}^{-1} {X}^\top {(\Sigma^k_N)}^{-1} {X} \right]
\right)}}{(2\pi)^{NQ/2} |\Sigma^k_Q|^{N/2} |\Sigma^k_N|^{Q/2}}.    
\end{split}
\end{equation}

Then, the NLL loss can be calculated as:
\begin{equation}\label{eq:loss_res}
  \begin{alignedat}{3}
  \mathcal{L}_{\text{NLL}} 
  &  = - \log p\left(R_t\right)   = - \log \sum_{k=1}^K \omega_t^k \cdot p^k_{MN}\left(R_t\,|\,{0},\Sigma^k_N,\Sigma^k_Q\right) \\
  &  = - \log \sum_{k=1}^K \exp   \Bigg( \log \omega^k_t - \frac{NQ}{2} \log (2\pi) 
   + N \log |\Sigma_Q^k|^{-\frac{1}{2}}  
       + Q \log |\Sigma_N^k|^{-\frac{1}{2}}  
   - \frac{1}{2} \Tr \left[ {(\Sigma_Q^{k})}^{-1} {R_t}^\top {(\Sigma_N^k)}^{-1} {R_t} \right] \Bigg).
  \end{alignedat}
\end{equation}

\begin{figure}
    \centering
\begin{adjustbox}{width=0.6\textwidth}
\begin{tikzpicture}[x=0.75pt,y=0.75pt,yscale=1,xscale=1, node distance=1.5cm, every node/.style={fill=white, font=\sffamily}, align=center]
    \node [text width=8cm, anchor=west, align=left] () at (-50,35) {(a) Training with conventional MSE loss};
    \node [text width=1cm] () at (170,10) {$f_M$};
    \node [text width=1cm] () at (170,-140) {$f_M$};
    \node [text width=8cm, anchor=west, align=left] () at (-50,-45) {(b) Training with NLL loss};

    \node (start_a)[Start]{$X_t$\\$[N\times P]$};
    \node (base_a)[basemodel, right of=start_a, xshift=1.5cm]{Base Model};
    \node (M_a)[rect, right of=base_a, xshift=1.5cm]{$M_t$\\$[N\times Q]$};
    \node (L_a)[loss, right of=M_a, xshift=1.5cm]{$\Lb_{\mathrm{MSE}}$};
    
    \path [draw,->] (start_a) edge (base_a);
    \path [draw,->] (base_a) edge (M_a);
    \path [draw,->] (M_a) edge (L_a);
    
    \draw [dashed, thick, color=black] (-30,-30) -- (370,-30);
    
    \node [text width=1cm] () at (170,-70) {$f_\omega$};
    \node (start_b)[Start] at (0, -80) {$X_t$\\$[N\times P]$};
    \node (base_b)[basemodel, right of=start_b, xshift=1.5cm]{Base Model};
    \node (w_b)[rect, right of=base_b, xshift=1.5cm, text width=2.2cm]{$\omega_t^k$\\$[K]$};
    
    \node (M_b)[rect, below of=w_b, yshift=0cm, text width=2.2cm]{$M_t$\\$[N\times Q]$};
    
    \node (Ls_b)[rect, below of=M_b, yshift=0cm, text width=2.2cm]{$L_S^k$\\$[K\times N\times N]$};
    \node (Lt_b)[rect, below of=Ls_b, yshift=0cm, text width=2.2cm]{$L_Q^k$\\$[K\times Q\times Q]$};
    \node (Lstrr_b)[loss, below of=L_a, yshift=-1.1cm]{$\Lb_{\text{NLL}}$};
    
    \node (no)[left of=Lstrr_b, xshift=-.12cm]{};
    \node (no2)[left of=w_b, xshift=-.12cm]{};
    
    \path [draw,->] (start_b) edge (base_b);
    \path [draw,->] (base_b) edge (M_b);
    \draw[->] (base_b.east) -| (no2.east) -- (w_b.west);
    \draw[->] (w_b.east) -| (no.east) -- (Lstrr_b.west);
    \draw (Lt_b.east) -| (no.east) -- (Lstrr_b.west);
    \draw (M_b.350) -|  (no.east);
\end{tikzpicture}
    \end{adjustbox}
    \caption{The model training with (a) conventional MSE loss and with (b) the proposed method.}
    \label{fig:my_label}
\end{figure}

For ease of computation, we can further parameterize $\Sigma^k_N$ and $\Sigma^k_Q$ using the Cholesky factorization of the precision (inverse of covariance) matrices, $\Lambda^k_N$ and $\Lambda^k_Q$:
\begin{equation}\label{eq:simple}
\begin{split}
   &{(\Sigma^k_N)}^{-1} = \Lambda^k_S = L^k_N {L^k_N}^\top,\\
   &{(\Sigma^k_Q)}^{-1} = \Lambda^k_Q = {L^k_Q} {L^k_Q}^\top,\\
   & \log |\Sigma_N^k|^{-\frac{1}{2}}  = \log \left| \Lambda^k_N \right|^{\frac{1}{2}} = \sum_{n=1}^N \log [L^k_N]_{n,n},\\
   & \log |\Sigma_Q^k|^{-\frac{1}{2}}  = \log \left| \Lambda^k_Q \right|^{\frac{1}{2}} = \sum_{q=1}^Q \log [L^k_Q]_{q,q}.\\
\end{split}
\end{equation}
where $L^k_N$ and $L_Q^k$ are the $k$-th spatial and temporal Cholesky factors (i.e., lower triangular matrices with positive diagonal entries), respectively.
The advantage of using the parameterization in Eq.~\eqref{eq:simple} is that we can avoid inverse operation during training, since it is hard to guarantee the \textit{invertibility} of a matrix during the training process. During implementation, we found that directly parameterizing with covariance matrices can result in divergence of the loss function, gradient exploding, and singular (non-invertible) covariance matrix. We resolved these issues by parameterizing the precision matrix. Since we do not have to invert the matrix during training, the training process is much more stable than parameterizing the covariance matrix using Cholesky factorization. Also, initializing the Cholesky factors as diagonal matrices helped stabilize the training process as well. Furthermore, we can avoid computing inverse operations during the training and efficiently compute the determinant by summing the logarithms of diagonal entries in the Cholesky factor.

The trace term can be simplified into computing the square of Frobenius Norm of $Z_t^k = {(L^k_N)}^\top R_t^k {L^k_Q}$ as
\begin{equation}\label{eq:simple2}
  \begin{split}
  \Tr\left[ \Lambda^k_Q {(R_t^k)}^\top \Lambda^k_N R_t^k \right]& 
  = \Tr\left[ {L^k_Q} {(L^k_Q)}^\top {(R_t^k)}^\top {L^k_N} {(L^k_N)}^\top R_t^k \right]  \\
   &  = \Tr\left[Z_t^k {(Z_t^k)}^\top \right]  = \|Z_t^k\|_F^2.
\end{split}
\end{equation}

Finally, Eq.~\eqref{eq:loss_res} can be simplified into
\begin{equation}\label{eq:nll_update1}
  \begin{alignedat}{3}
  &  \mathcal{L}_{\text{NLL}} = 
  & - \log \sum_{k=1}^K \exp && \Big( \log \omega^k_t - \frac{NQ}{2} \log (2\pi)  - \frac{1}{2} \|Z_t^k\|_F^2 
   + N \sum_{q=1}^Q \log [L^k_Q]_{q,q} +N\sum_{n=1}^N \log [L^k_N]_{n,n} \Big). 
  \end{alignedat}
\end{equation}

Note the term inside the \textit{exponential function} may become very large (either negative or positive), and this will make it infeasible to compute the log-likelihood. To avoid numerical problems in the training process, we use the Log-Sum-Exp trick $\log \sum_{k=1}^K \exp(z_k) = z^* + \log \sum_{k=1}^K \exp(z_k - z^*)$ with $z^* = \max \{ z_1, \cdots, z_K \}$. 

Finally, we can train a given baseline model based on Eq.~\eqref{eq:loss} by using both conventional MSE loss ($\mathcal{L}_{\text{MSE}}$) as well as the NLL loss ($\mathcal{L}_{\text{NLL}}$). We jointly train $f_{M}$, $f_{\omega}$, $L_Q^k$, and $L^k_N$ by minimizing the total loss ($\mathcal{L}_{\text{DynMix}}=\mathcal{L}_{\text{MSE}} + \rho \mathcal{L}_{\text{NLL}}$). Both $f_{M}$ and $f_{\omega}$ are represented as a neural network. We slightly change the last layer of the baseline model to output a hidden representation of the given input, and we use two separate Multi-layer Perceptron (MLP) for $f_{M}$ and $f_{\omega}$ as shown in Figure \ref{fig:my_label}. In other words, the baseline model is a shared layer for $f_{M}$ and $f_{\omega}$ to calculate a hidden representation of the input, and the last MLP is the function-specific module to calculate the desired output. We do not have an activation function at the last layer of MLP for $f_{M}$ since we generally use z-score normalization before feeding the input to the model, but we apply the softmax function at the last layer of MLP for $f_{\omega}$ to force it to be summed to 1. The gradient updates of $L_Q^k$ and $L^k_N$ are only applied to the ``lower'' part of the matrices, and the values in the upper parts remain as 0 during the whole process. In parameter initialization, we initialized both $L_Q^k$ and $L_N^k$ as a diagonal matrix, which was empirically more stable than random initialization.

\subsection{Discussion}

{

\begin{table}[]
    \centering
    \begin{tabular}{c|c|c|c}
    \toprule\toprule
        Assumed distribution  &  Time complexity  & spatiotemporal correlation & dynamic \\\midrule\midrule
        $\mathcal{N} (0, \sigma^2 I)$ (MSE) &  $O(NQ)$ & \xmark  & \xmark \\\midrule
        $\mathcal{N} (0, \Sigma)$ &  $O(N^3 Q^3)$ & \cmark  & \xmark \\\midrule
        $\mathcal{N} (0, \Sigma_Q \otimes \Sigma_N)$ &  $O(N^3 + Q^3)$ & \cmark  & \xmark \\\midrule
        \makecell{$ \sum_{k=1}^K \omega_t^k \mathcal{N} (0, \Sigma_Q \otimes \Sigma_N)$\\(proposed)} &  $O(K(N^3 + Q^3) )$ & \cmark  & \cmark \\
         \bottomrule\bottomrule
    \end{tabular}
\caption{Computational efficiency comparison of various distribution assumptions.}
\label{tab:comp}
\end{table}

\subsubsection{Computational efficiency}

Table \ref{tab:comp} shows different assumptions for output distribution as well as the time complexity, whether the model accounts for spatiotemporal correlation, and whether the model can capture dynamic distribution. 
Using MSE ($\mathcal{N} (0, \sigma^2 I)$) shows the lowest time complexity; however, it can neither model the spatiotemporal correlation nor the dynamic characteristics of the distribution. 
In contrast, using multivariate Gaussian with full covariance ($\mathcal{N} (0, \Sigma)$) can capture the spatiotemporal correlation, however, it has a significant cubic time complexity with respect to both spatial dimension ($N$) and temporal dimension ($Q$), resulting in $O(N^3 Q^3)$. This is not realistically feasible to implement.
The third option, using a decomposed covariance structure ($\mathcal{N}(0, \Sigma_Q \otimes \Sigma_N)$), strikes a balance between capturing spatiotemporal correlations and managing computational complexity. By assuming that the whole spatiotemporal covariance can be decomposed into spatial ($\Sigma_N$) and temporal ($\Sigma_Q$) covariances, this model reduces the overall computational cost to $O(N^3 + Q^3)$. This significantly mitigates the computational burden compared to the full covariance approach, making it more feasible for larger datasets or more frequent analysis without sacrificing the ability to model the interactions between space and time. However, this approach still cannot capture the dynamic characteristics since the assumed distribution does not have any time-dependent component in it.
Finally, the proposed method (Dynamic Mixture) assumes each component Gaussian has its own spatiotemporal covariance structure represented as $\sum_{k=1}^K \omega_t^k \mathcal{N}(0, \Sigma_Q \otimes \Sigma_N)$. This approach allows for modeling dynamic distribution by introducing weights ($\omega_t^k$) that change with respect to the input ($X_t$), thereby capturing both the spatiotemporal correlations and the dynamic characteristics of the distribution. The computational complexity for this method is $O(K(N^3 + Q^3))$, where $K$ is the number of mixture components. While this increases the computational demands relative to the simpler decomposed covariance structure, it offers a more detailed and flexible modeling of the data dynamics, making it suitable for applications where temporal changes in the data structure are significant and need to be accurately reflected in the model. This method represents a comprehensive approach, providing a robust framework for analyzing complex spatiotemporal data with varying dynamics.

}

\subsubsection{Comparison with Generalized Least Squares Loss}

The proposed model aims to effectively deal with spatiotemporally correlated errors by modeling the time-varying distribution of the spatiotemporal matrix process. By utilizing a probabilistic sequence-to-sequence forecasting model, we are able to characterize the error process with a predictable distribution. This enables us to predict both the mean and the error distribution, providing a more comprehensive forecasting model.

The proposed method, particularly when $K=1$, is conceptually related to the generalized least squares (GLS) loss function \citep{kariya2004generalized}. 
{
Similar to GLS, the proposed method accounts for heteroscedasticity and correlated errors by incorporating the covariance structure into the loss function. Previous studies on GLS have also explored the use of the Kronecker product to decompose the covariance matrix for computational efficiency \citep{fausett1994large, karimi2018global, marco2019least}. 
}
However, while our method shares conceptual similarities with GLS, there are significant differences in both theoretical formulation and practical implementation. These differences include how the covariance matrix is handled, the role of determinant regularization, and the ability to capture multimodal distributions through a mixture model framework.
{
First, unlike conventional GLS, which rigorously selects the dispersion matrix (or precision matrix) either by relying on prior knowledge (i.e., predefined covariance) or by iteratively estimating it \citep{yang2019estimating}, the proposed method \textbf{learns the covariance matrix directly during the model training process}. This approach allows the covariance matrix to be dynamically adapted in a data-driven manner through a training process. By jointly optimizing the covariance matrix as part of the learning process, the proposed method enhances the model's flexibility and robustness in handling complex, high-dimensional data.
}
Secondly, our method begins with the maximum likelihood estimation of the (mixture of multivariate Gaussian) distribution and consequently incorporates the \textbf{determinant of the covariance matrix} (or precision matrix) in the loss function. This inclusion does not occur in GLS since the covariance is predetermined and its determinant does not change during training. The determinant term is particularly crucial when the covariance is being updated during the training process as it acts as a \textbf{natural regularizer} for the learned covariance matrix. Essentially, our method finds the GLS solution while simultaneously regularizing (minimizing) the determinant of the covariance matrix, which provides a more suitable loss function for training neural networks. 
Finally, the proposed model can be extended to a mixture model when $K>1$. This flexibility allows the proposed method to \textbf{capture more complex and dynamic characteristics of the data}, which are often encountered in practical applications. The scalability of the method to $K>1$ enables the handling of multimodal distributions and adapts to changes in data dynamics over time, providing a robust framework for analyzing intricate spatiotemporal datasets.

\subsubsection{Multimodality}

One of the key advantages of the proposed model is its ability to handle errors with a multimodal structure. By utilizing a mixture model, we are able to model the multimodal distribution and offer interpretability for the different modes present in the errors. This is particularly useful in situations where there are multiple potential outcomes or patterns (i.e., multimodality) present in the errors. This feature allows the proposed model to be more robust to the complex and diverse nature of spatiotemporal data.

\subsubsection{Connection with Mixture Density Network}

The proposed model also stands out for being a special case of a matrix-variate Mixture Density Network. This architecture allows us to handle the complexities of spatiotemporal data by leveraging the matrix-variate nature of the data, resulting in a more nuanced and accurate forecasting model. The proposed model does not output the Cholesky factors (i.e., $L_N^k$ and $L_Q^k$), while previous studies using a mixture density network either output the variances by assuming a diagonal covariance matrix or output a full covariance matrix from the base line model. The proposed approach can avoid over-fitting and offer interpretability, which is a crucial aspect in many real-world applications. 

Overall, our proposed model offers a powerful approach for dealing with spatiotemporally correlated errors in a probabilistic and interpretable manner. It provides a comprehensive forecasting model that can effectively handle multimodal errors and the complexities of spatiotemporal data, while avoiding over-fitting and providing interpretability. This makes the proposed model a valuable tool for various real-world applications that deal with spatiotemporal data.

\section{Experiment and Results}\label{sec:result}

\subsection{Experimental Setup}
We conduct experiments to verify the performance of the proposed model. The datasets we use include two widely-used traffic speed datasets \citep{li2017diffusion,yu2017spatio, wu2019graph}, \textbf{PEMS-BAY-2017-SPEED} and \textbf{METR-LA}, and two new datasets, \textbf{PEMS-BAY-2022-FLOW} and \textbf{PEMS-BAY-2022-SPEED}\footnote{\url{https://github.com/benchoi93/PeMS-BAY-2022}}.
PEMS-BAY-2017-SPEED, PEMS-BAY-2022-SPEED, and PEMS-BAY-2022-FLOW were collected from the California Transportation Agencies (CalTrans) Performance Measurement System (PeMS) \citep{chen2001freeway}. They are collected from 325 sensors installed in the Bay Area with observations of 6 months. PEMS-BAY-2017-SPEED uses speed data ranging from January 1st to May 31st in 2017, PEMS-BAY-2022-SPEED uses speed data ranging from January 1st to June 30th in 2022, and PEMS-BAY-2022-Flow uses flow data ranging from January 1st to June 30th in 2022. METR-LA was collected from the highways of Los Angeles County \citep{jagadish2014big}. It is collected from 207 sensors with observations of 4 months of data ranging from March 1 to June 30, 2012. All datasets use a 5-min resolution for the observations.

We choose Root Mean Square Error (RMSE), Mean Absolute Percentage Error (MAPE), and Mean Absolute Error (MAE) of speed prediction at 15, 30, 45, and 60 minutes ahead as key metrics to evaluate forecasting accuracy, corresponding to 3-, 6-, 9-, and 12-step-ahead prediction, respectively.

The proposed method can work as an add-on to any type of traffic forecasting model based on deep learning. We use Graph Wavenet (\textbf{GWN}) \citep{wu2019graph} and Spatio-Temporal Graph Convolutional Networks (\textbf{STGCN}) \citep{yu2017spatio} as our baseline models. Following the conventional configurations, we aim to predict 12 steps (i.e., one hour in the future) based on 12-step observation (i.e., observations from one hour in the past). 

{
We first run Bayesian Optimization on the baseline model without using the proposed loss. This step ensures that the baseline model is performing at its best. The hyperparameters optimized during this stage include:
\begin{itemize}
    \item Learning rate: $[ 10^{-5} , 10^{-1} ]$
    \item Weight decay: $[0.9, 0.999]$
    \item Number of hidden neurons per layer: $[16, 128]$
    \item Dropout: $[0.1, 0.5]$
\end{itemize}

Once we obtain the best hyperparameters for the baseline model, we introduce the proposed loss function. At this stage, we chose to try all $K$'s from 1 to 5 to show the performance of the model with different $K$.
}

All models are tested based on the official code implementations provided by the authors of two previous works\footnote{\url{https://github.com/nnzhan/Graph-WaveNet}}\footnote{\url{https://github.com/VeritasYin/STGCN_IJCAI-18}}. All models are tested based on different random seeds ranging from 1 to 5, and the results show the average performance of the five runs.

\subsection{Results}
Table \ref{tab:result} shows the summary of the aggregated results. First of all, in most cases, the proposed method can improve the prediction performance of the trained models in terms of RMSE, MAPE, and MAE. 
In the PEMS-BAY-2017-SPEED, we observe that while the GWN model provides a baseline performance, its augmentation with $\mathcal{L}_{\text{DynMix}}$ results in marked improvements in all metrics, particularly RMSE, MAPE, and MAE. This improvement is more pronounced in longer forecasting horizons, highlighting the model's enhanced capability to handle the complexities of traffic dynamics over extended periods. The $\mathcal{L}_{\text{DynMix}}$(3) and $\mathcal{L}_{\text{DynMix}}$(4) variants show the best performance, especially in reducing MAPE, which is crucial in congested traffic scenarios.
Similarly, for the METR-LA Dataset, the addition of $\mathcal{L}_{\text{DynMix}}$ to the GWN model leads to significant improvements, particularly in RMSE and MAE at the 60-minute forecasting horizon. The most consistent enhancements across all metrics are observed with the $\mathcal{L}_{\text{DynMix}}$(2) variant, suggesting its effectiveness in this specific traffic forecasting context.
The results for the PEMS-BAY-2022-FLOW Dataset also reflect this trend. The GWN model, when augmented with $\mathcal{L}_{\text{DynMix}}$, especially in the $\mathcal{L}_{\text{DynMix}}$(4) variant, shows substantial improvements in prediction accuracy, particularly for congested traffic states as indicated by the MAPE metric. The STGCN model benefits significantly from the $\mathcal{L}_{\text{DynMix}}$ enhancement, with the $\mathcal{L}_{\text{DynMix}}$(2) variant showing the most notable improvements in RMSE and MAPE.
Furthermore, in the PEMS-BAY-2022-SPEED Dataset, the integration of $\mathcal{L}_{\text{DynMix}}$ with the GWN model leads to notable improvements in forecasting accuracy, especially in longer horizons. The $\mathcal{L}_{\text{DynMix}}$(1) and $\mathcal{L}_{\text{DynMix}}$(2) variants emerge as the most effective in capturing the spatiotemporal dynamics of traffic speed. The STGCN model also shows significant enhancements in performance, particularly in MAPE and MAE, with the best results seen in the $\mathcal{L}_{\text{DynMix}}$(5) variant.

Overall, these results underscore the effectiveness of the $\mathcal{L}_{\text{DynMix}}$ loss function in improving traffic forecasting models. 
It is notable that the performance improvement is greater for the longer forecasting horizon (60~min) than for the shorter forecasting horizon (15~min). This result is intuitive that we usually have stronger spatiotemporal correlation at longer forecasting horizons than that at shorter forecasting horizons. 
A key aspect of the results is the notable improvement in the performance of the Mean Absolute Percentage Error (MAPE) for speed datasets. MAPE is known for penalizing large errors especially in congested traffic states where speeds are low. The most correlated errors in traffic forecasting occur during congested states, as suggested by \citet{li2022estimate}. In contrast, errors in a free-flow state generally follow an independent Gaussian distribution. This distinction is critical for understanding the nature of traffic forecasting errors and tailoring models accordingly. The effectiveness of the proposed method in improving MAPE underscores its utility in accurately predicting traffic conditions in challenging cases.

{
It is important to note that the performance of different variants of the proposed model is dependent on the specific characteristics and complexities of each dataset as well as the choice of the baseline model. Each dataset may have unique statistical properties, noise characteristics, and underlying dynamics. Some datasets may exhibit multimodal distributions or significant temporal dynamics that are better captured by models with a higher $K$, allowing for more complex mixture models to better represent the data. However, higher $K$ values may lead to overfitting in datasets with more uniform statistical properties or simpler dynamics, where additional components do not contribute meaningful information. Thus, the optimal configuration of $K$ is inherently data-dependent, and its selection is critical for achieving the best performance.
}

\begin{table*}[!t]
  \centering
  \caption{Summary of results showing the performance metrics (RMSE, MAPE, and MAE) at 15-, 30-, 45-, and 60-min prediction. The shown values are the average of 5 runs. The best result in each column is bolded with a star mark}
  \label{tab:result}
  \resizebox{\textwidth}{!}{
\begin{tabular}{c|l|cccc|cccc|cccc}\toprule\toprule
\multirow{2}{*}{Data}  &  \multirow{2}{*}{Model}&   \multicolumn{4}{c}{RMSE}  & \multicolumn{4}{c}{MAPE}  &  \multicolumn{4}{c}{MAE} \\\cmidrule{3-14}
 & & 15~min & 30~min & 45~min & 60~min & 15~min & 30~min & 45~min & 60~min & 15~min & 30~min & 45~min & 60~min \\\midrule\midrule
\multirow{12}{*}{\shortstack{PEMS-BAY\\ 2017\\ Speed}} 
&  GWN  & 
2.64&3.46&3.88&4.16&2.93&3.99&4.61&5.07&1.40&1.77&1.98&2.14
\\
 &  \quad+$\mathcal{L}_{\text{DynMix}}$(1)  &
\best{2.58}&3.40&\best{3.77}&\best{4.00}&2.96&3.98&4.45&4.79&1.35&1.70&1.87&1.99
\\
&  \quad+$\mathcal{L}_{\text{DynMix}}$(2)  & 
\best{2.58}&3.41&3.81&4.06&\best{2.81}&3.81&4.36&4.70&\best{1.34}&\best{1.68}&1.87&2.00
\\
 &  \quad+$\mathcal{L}_{\text{DynMix}}$(3)  &
2.57&\best{3.39}&\best{3.77}&\best{4.00}&2.83&\best{3.75}&4.29&4.64&\best{1.34}&\best{1.68}&\best{1.86}&\best{1.98}
 \\
 &  \quad+$\mathcal{L}_{\text{DynMix}}$(4)  & 
2.57&3.40&3.81&4.06&2.87&3.82&4.33&4.71&\best{1.34}&1.69&1.88&2.00
 \\
 &  \quad+$\mathcal{L}_{\text{DynMix}}$(5)  & 
2.57&3.40&3.81&4.06&2.78&3.72&\best{4.24}&\best{4.61}&\best{1.34}&1.69&1.89&2.02
  \\
 \cmidrule{2-14}
 &  STGCN  &  3.24 & 3.94 & 4.35 & 4.66 & 4.36 & 5.40 & 6.08 & 6.59 & 1.86 & 2.15 & 2.35 & \best{2.51}  \\
 &  \quad+$\mathcal{L}_{\text{DynMix}}$(1)  &2.98 & 3.84 & 4.31 & 4.68 & 3.90 & 5.12 & 5.82 & 6.45 & 1.64 & 2.05 & 2.32 & 2.55  \\
 &  \quad+$\mathcal{L}_{\text{DynMix}}$(2)  &\best{2.87} & \best{3.74} & \best{4.21} & \best{4.58} & 3.60 & 4.88 & \best{5.66} & \best{6.36} & \best{1.55} & \best{1.99} & \best{2.28} & 2.52 \\
 &  \quad+$\mathcal{L}_{\text{DynMix}}$(3)  &2.98 & 3.86 & 4.34 & 4.72 & 3.91 & 5.22 & 6.02 & 6.78 & 1.61 & 2.08 & 2.38 & 2.63  \\
 &  \quad+$\mathcal{L}_{\text{DynMix}}$(4)  & 2.88 & 3.76 & 4.25 & 4.63 &\best{ 3.59} & \best{4.87} & \best{5.66} & 6.40 & 1.56 & 2.02 & 2.34 & 2.59  \\
 &  \quad+$\mathcal{L}_{\text{DynMix}}$(5)  & 2.95 & 3.78 & 4.31 & 4.62 & 3.95 & 5.11 & 6.04 & 6.54 & 1.63 & 2.05 & 2.37 & 2.59  \\
 \midrule\midrule
\multirow{12}{*}{\shortstack{METR-LA}} 
&  GWN  & 5.14 & 6.11 & 6.73 & 7.22 & 7.59 & 9.22 & 10.34 & 11.21 & 2.98 & 3.53 & 3.92 & 4.24 
\\
 &  \quad+$\mathcal{L}_{\text{DynMix}}$(1)  &4.89 & 5.84 & 6.37 & 6.74 & \best{7.15} & 8.82 & 9.94 & 10.76 & 2.83 & 3.29 & 3.60 & 3.81 
\\
 &  \quad+$\mathcal{L}_{\text{DynMix}}$(2)  & \best{4.87} & \best{5.77} & \best{6.32} & \best{6.69} & 7.27 & 8.73 & \best{9.71} & \best{10.38} & 2.83 & \best{3.25} & \best{3.54} & \best{3.77}
\\
 &  \quad+$\mathcal{L}_{\text{DynMix}}$(3)  & 4.93 & 5.92 & 6.57 & 6.96 & 7.17 & \best{8.71} & 9.80 & 10.56 & 2.92 & 3.45 & 3.84 & 4.07 
 \\
 &  \quad+$\mathcal{L}_{\text{DynMix}}$(4)  & 4.88 & 5.78 & \best{6.32} & \best{6.69} & 7.30 & 9.02 & 10.02 & 10.90 & \best{2.81} & 3.27 & 3.56 & 3.79 
 \\
 &  \quad+$\mathcal{L}_{\text{DynMix}}$(5)  & \best{4.87} & 5.79 & 6.36 & 6.77 & 7.46 & 8.95 & 9.99 & 10.74 & 2.86 & 3.33 & 3.65 & 3.91 
  \\
 \cmidrule{2-14}
 &  STGCN  &  5.94 & 6.78 & 7.33 & 7.77 & 9.31 & 10.60 & 11.54 & 12.37 & 3.69 & 4.16 & 4.51 & 4.82 \\
 &  \quad+$\mathcal{L}_{\text{DynMix}}$(1)  & 5.42 & 6.35 & 6.93 & 7.38 & 8.93 & 10.76 & 12.04 & 13.13 & 3.30 & 3.85 & 4.26 & 4.59 \\
 &  \quad+$\mathcal{L}_{\text{DynMix}}$(2)  & \best{5.18} & \best{6.02} & \best{6.54} & \best{6.91} & \best{8.02}&\best{ 9.65} & \best{10.77} & \best{11.67} & \best{3.10}&\best{ 3.55} &\best{3.86} & \best{4.13}  \\
 &  \quad+$\mathcal{L}_{\text{DynMix}}$(3)  & 5.27 & 6.13 & 6.64 & 7.02 & 8.22 & 9.85 & 10.90 & 11.71 & 3.19 & 3.67 & 4.00 & 4.26  \\
 &  \quad+$\mathcal{L}_{\text{DynMix}}$(4)  & 5.26 & 6.09 & 6.59 & 6.94 & 8.45 & 10.07 & 11.12 & 11.96 & 3.22 & 3.68 & 3.99 & 4.25  \\
 &  \quad+$\mathcal{L}_{\text{DynMix}}$(5)  & 5.29 & 6.13 & 6.62 & 6.98 & 8.49 & 10.12 & 11.16 & 12.01 & 3.25 & 3.71 & 4.03 & 4.29  \\
 \bottomrule\bottomrule 
\multirow{12}{*}{\shortstack{PEMS-BAY\\ 2022\\ Flow}} 
&  GWN  & 25.34 & 28.70 & 31.16 & 33.90 & 13.74 & 14.80 & 17.12 & 20.73 & 16.90 & 19.26 & 21.18 & 23.52 \\
 &  \quad+$\mathcal{L}_{\text{DynMix}}$(1)  & 30.12 & 37.81 & 42.86 & 45.60 & 13.00 & 15.64 & 17.99 & 18.67 & 21.31 & 27.44 & 31.51 & 33.49 \\
 &  \quad+$\mathcal{L}_{\text{DynMix}}$(2)  & 24.14 & 26.92 & 29.05 & 30.73 & 12.27 & 14.87 & 17.18 & 18.20 & 15.78 & 17.66 & 19.31 & \best{20.56} \\
 &  \quad+$\mathcal{L}_{\text{DynMix}}$(3)  & \best{24.02} & \best{27.11} & 29.23 & 30.83 & 11.86 & 14.21 & 16.77 & 17.59 & 15.67 & 17.82 & 19.50 & 20.66 \\
 &  \quad+$\mathcal{L}_{\text{DynMix}}$(4)  & 23.86 & 26.77 & \best{29.00} & \best{31.04} & 12.10 & 13.53 & 15.06 & \best{16.03} & \best{15.51} & \best{17.45} & \best{19.12} & 20.69 \\
 &  \quad+$\mathcal{L}_{\text{DynMix}}$(5)  & 24.32 & 27.17 & 29.48 & 31.06 & \best{11.78} & \best{13.28} & \best{14.45} & 16.07 & 15.95 & 17.78 & 19.46 & 20.64 \\
 \cmidrule{2-14}
 &  STGCN  &  39.66 & 41.69 & 43.40 & 45.52 & 35.08 & 35.57 & 35.42 & 35.21 & 30.19 & 31.36 & 32.43 & 33.80  \\
 &  \quad+$\mathcal{L}_{\text{DynMix}}$(1)  & 31.78 & 38.71 & 44.04 & 47.44 & 15.56 & 17.39 & 19.18 & 20.66 & 22.82 & 27.99 & 32.15 & 34.87 \\
 &  \quad+$\mathcal{L}_{\text{DynMix}}$(2)  & 28.50 &\best{ 31.35} & \best{33.97} & \best{36.12}& \best{15.18} & \best{16.03} & \best{17.34} & \best{18.59} & \best{20.10} & \best{21.85} & \best{23.82} & \best{25.39} \\
 &  \quad+$\mathcal{L}_{\text{DynMix}}$(3)  & 29.26 & 32.03 & 34.45 & 36.62 & \best{15.18} & 16.35 & 17.54 & 18.78 & 20.65 & 22.45 & 24.21 & 25.84  \\
 &  \quad+$\mathcal{L}_{\text{DynMix}}$(4)  & \best{28.44} & \best{31.35} & 34.02 & 36.22 & 15.21 & 16.32 & 17.37 & 18.70 & 20.08 & 21.95 & 23.87 & 25.50  \\
 &  \quad+$\mathcal{L}_{\text{DynMix}}$(5)  & 28.51 & 31.51 & 34.09 & 36.39 & 15.53 & 16.70 & 17.58 & 18.83 & 20.11 & 22.08 & 23.86 & 25.58  \\
 \bottomrule\bottomrule 
\multirow{12}{*}{\shortstack{PEMS-BAY\\ 2022\\ Speed}} 
&  GWN  & 2.28 & 2.89 & 3.22 & 3.47 & 2.32 & 2.96 & 3.37 & 3.70 & 1.17 & 1.41 & 1.58 & 1.72 \\
 &  \quad+$\mathcal{L}_{\text{DynMix}}$(1)  & \best{2.10} & \best{2.70} & \best{3.00} & 3.18 & 2.05 & 2.66 & 3.01 & 3.23 & \best{1.04} & \best{1.28} & \best{1.42} & \best{1.52} \\
 &  \quad+$\mathcal{L}_{\text{DynMix}}$(2)  & \best{2.10} & \best{2.70} & 2.99 & \best{3.17} & 2.03 & 2.64 & 3.01 & 3.26 & \best{1.04} & 1.29 & 1.43 & 1.52 \\
 &  \quad+$\mathcal{L}_{\text{DynMix}}$(3)  & \best{2.10} & 2.71 &\best{ 3.00} & 3.18 & \best{2.01} & \best{2.63} & \best{2.98} & \best{3.21} & \best{1.04} & 1.30 & 1.44 & 1.54 \\
 &  \quad+$\mathcal{L}_{\text{DynMix}}$(4)  & \best{2.10} & 2.72 & 3.02 & 3.20 & 2.02 & 2.70 & 3.06 & 3.29 & \best{1.04} & 1.30 & 1.44 & 1.54 \\
 &  \quad+$\mathcal{L}_{\text{DynMix}}$(5)  & \best{2.10} & 2.72 & 3.02 & 3.18 & 2.02 & 2.68 & 3.05 & 3.26 & 1.05 & 1.31 & 1.45 & 1.55 \\
 \cmidrule{2-14}
 &  STGCN  & 2.87 & 3.33 & 3.63 & 3.87 & 3.14 & 3.55 & 3.88 & 4.17 & 1.62 & 1.81 & 1.96 & 2.10 \\
 &  \quad+$\mathcal{L}_{\text{DynMix}}$(1)  &  2.38 & 2.96 & 3.27 & 3.49 & 2.53 & 3.16 & 3.54 & 3.82 & 1.27 & 1.53 & 1.70 & 1.83  \\
 &  \quad+$\mathcal{L}_{\text{DynMix}}$(2)  & \best{2.22} & \best{2.83} & 3.15 & 3.36 & {2.33} & 3.04 & 3.44 & 3.72 & 1.16 & 1.45 & 1.62 & 1.75  \\
 &  \quad+$\mathcal{L}_{\text{DynMix}}$(3)  & \best{2.22} & 2.84 & 3.16 & 3.37 & 2.32 & 3.03 & 3.44 & 3.71 & 1.16 & 1.44 & 1.61 & 1.74  \\
 &  \quad+$\mathcal{L}_{\text{DynMix}}$(4)  &  2.23 & 2.87 & 3.19 & 3.40 & \best{2.30} & \best{3.02} & 3.44 & 3.71 & \best{1.15} & 1.43 & 1.60 & 1.73 \\
 &  \quad+$\mathcal{L}_{\text{DynMix}}$(5)  & 2.23 & \best{2.83} & \best{3.13} & \best{3.33} & 2.35 & \best{3.02} & \best{3.41} & \best{3.66}  & \best{1.15} & \best{1.42} & \best{1.58} & \best{1.71} \\
 \bottomrule\bottomrule 
\end{tabular}
}
\end{table*}
\clearpage

\begin{figure}[!htb]
    \centering
    \includegraphics[width=0.54\textwidth]{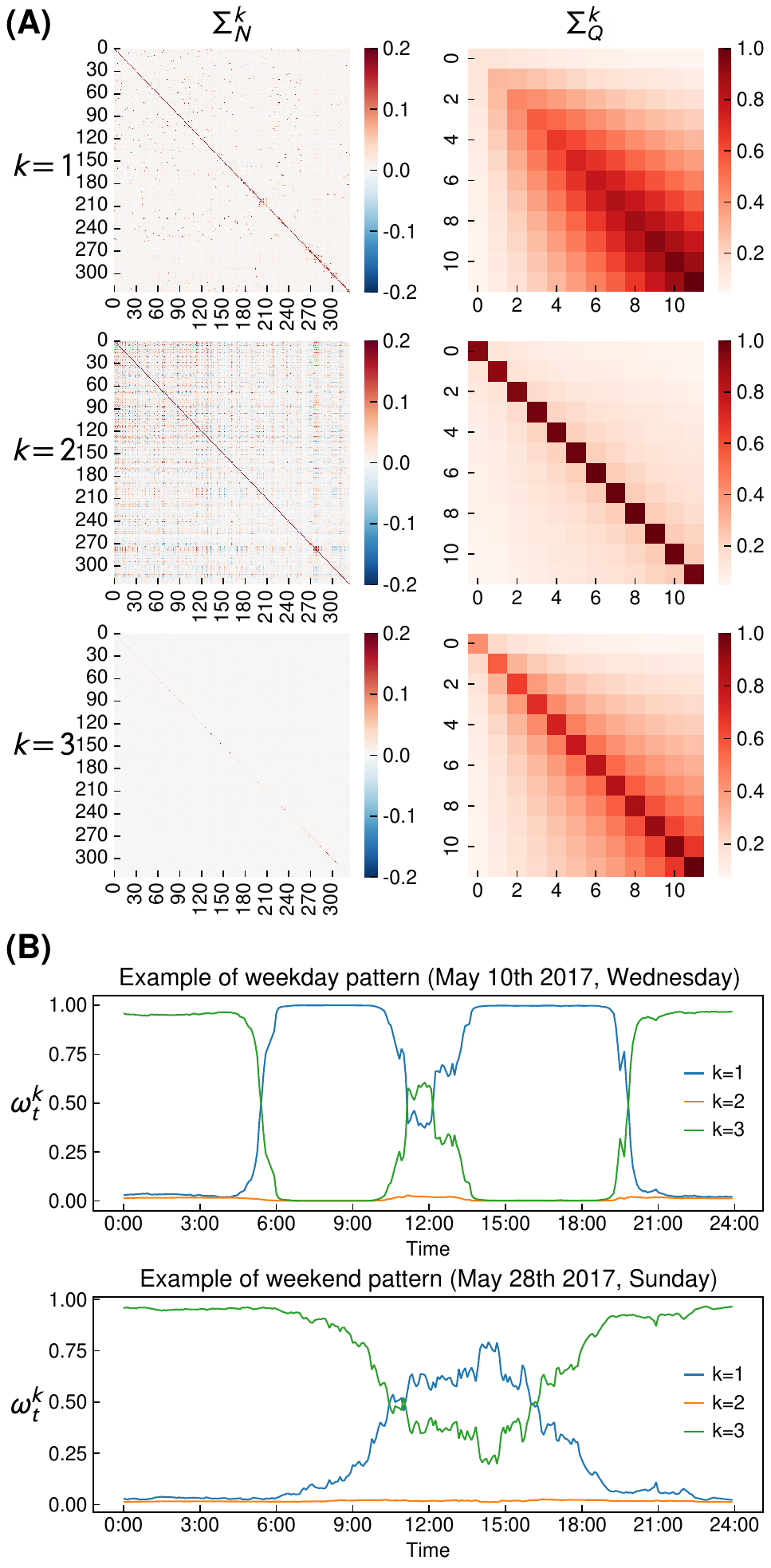}
    \caption{\textbf{(A)} Learned spatial ($\Sigma_N^k$) and temporal ($\Sigma_Q^k$) covariance matrices with $K=3$ using GWN as base model on PEMS-BAY-2017-SPEED dataset. For better visualization, we normalized the temporal covariance matrix by dividing the temporal covariance matrices by the maximum value of the diagonal entries in temporal covariance matrices and multiplied the same value to the spatial covariance matrices, since the Kronecker product has the scale-invariant property, i.e., $A\otimes B = \left(\nu A\right) \otimes \left(\frac{1}{\nu} B\right)$.
    \textbf{(B)}~Examples of patterns of the mixture weight $\omega_t^k$. There were two distinct patterns for 35 days in the testing dataset, which could be categorized into weekday-pattern and weekend-pattern. The representative cases for each category are shown.
    }
    \label{fig:result_summary}
\end{figure}

One advantage of using the proposed method is that representing the error distribution with the dynamic mixture distribution offers an interpretable framework for traffic speed forecasting. We selected $K=3$ with MSE base loss on PEMS-BAY dataset using GWN model for this test. 
Figure \ref{fig:result_summary} (A) shows the learned spatial and temporal covariance matrices. 
We can observe the spatial correlation structure of the errors in each mixture component, and the spatial structure is clearer in the first component. The temporal covariance matrices show an increasing tendency of variance/covariance as the forecasting time horizon increases. This is intuitive since the variance of the forecasting increases at a longer forecasting time step, and we can verify that the proposed method enables learning this type of characteristic of traffic speed forecasting. 
%

Figure \ref{fig:result_summary} (B) shows the change of $\omega_t^k$ over different time-of-day. The x-axis is the time-of-day ranging from 0:00 to 23:55 with 5-minute-interval, and the y-axis is the value for $\omega_t^k$. It is important to note that the sum of $\omega_t^k$ should equal to 1 at each time-of-day. During examining this result, we found that there are two distinct patterns in 35 days of data in the testing dataset. Surprisingly and trivially, these two patterns can be classified as weekday and weekend patterns. The upper figure shows the representative example of weekday pattern from Wednesday, May 10th, 2017, while the lower figure shows the representative example of weekend pattern from Sunday, May 28th, 2017. In both cases, the pattern in the first mixture component ($k=1$) is represented by the blue line and the pattern in the third component ($k=3$) pattern is represented by the green line and are the main components, and the influence of the second mixture component is small in both cases. In the weekday pattern, the pattern in the first mixture component is mainly dominant during peak hour periods including both morning peak (from 6:00 to 9:00) and afternoon peak (from 15:00 to 19:00) and the pattern in the third mixture component is mainly dominant during night-time from 21:00 to 5:00. In the afternoon off-peak, a mixture of the two patterns is observed. When interpreting these results together with the results of Figure \ref{fig:result_summary} (A), it can be seen that the results in spatial and temporal covariance matrices of the first component show a strong spatial and temporal correlations, which can be easily interpreted as ``peak-hour pattern.'' On the other hand, in the third component, the temporal correlation is strong, but the spatial covariance is almost diagonal, meaning that the speed observations are spatially independent, which can be interpreted as ``off-peak pattern.'' We have a mixture of two patterns in the afternoon off-peak hours since the distribution cannot be represented as unimodal distribution since there may or may not be congestion in some occasion. As a result, a mixture of the peak-hour and off-peak patterns is suitable for this time period. The weekend pattern shows a different pattern from the weekday pattern in that, at night time, it is similar to the weekday pattern of the third mixture component dominant, but during the day time, it is represented as a mixture of the first and the third mixture component, similar to the weekday's afternoon off-peak hours. 



\begin{figure}[t]
    \centering
    \includegraphics[width=\linewidth]{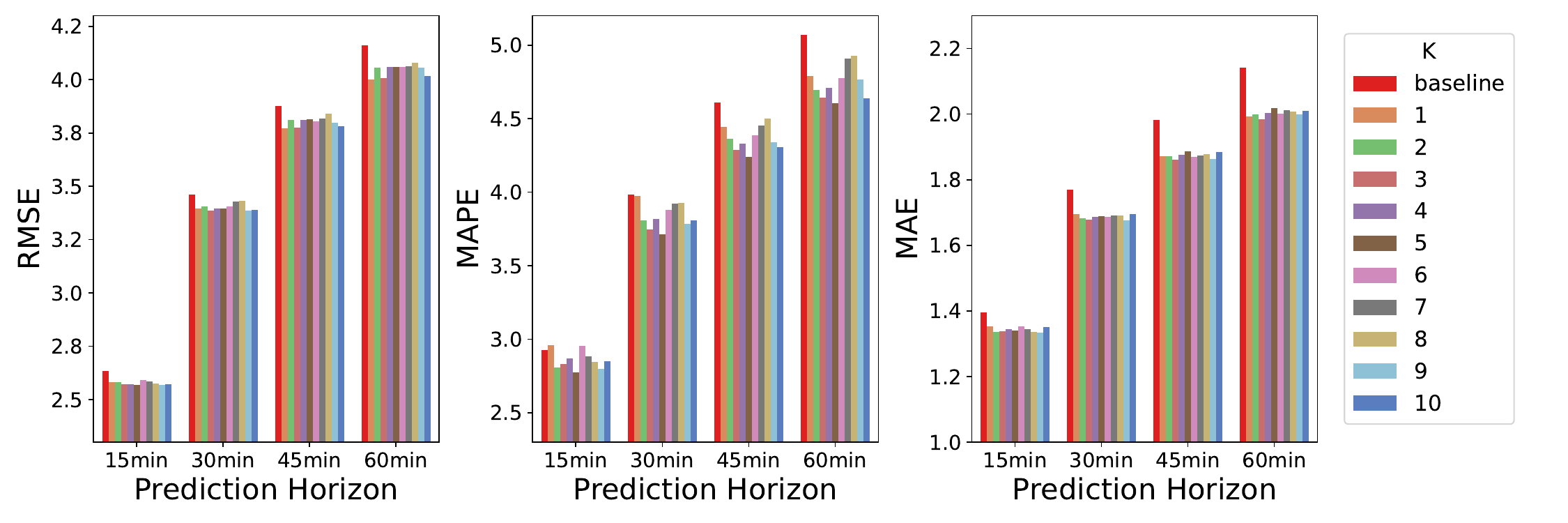}
    \caption{Ablation on prediction performance using the proposed loss function for $K=[1,...,10]$ using GWN as baseline model for PEMS-BAY 2017 dataset. } 
    \label{fig:abl_K}
\end{figure}

{
Figure \ref{fig:abl_K} illustrates one example of the impact of $K$ beyond the values presented in Table \ref{tab:result}. The proposed loss function consistently improves the model training performance across three metrics --- RMSE, MAPE, and MAE --- over various prediction horizons (15, 30, 45, and 60 minutes). Notably, the choice of $K$ directly influences the trade-off between the model complexity and robustness. Lower $K$ values generally provide a balance between simplicity and performance improvements, while higher $K$ allows the model to learn more complex patterns but risk overfitting. Theoretically, a mixture of Gaussian distributions can approximate any distribution; however, in practice, using a higher number of mixture components may lead to overfitting by capturing noise instead of meaningful patterns in the data \citep{chen2023probabilistic}.
}

\begin{figure}
    \centering
    \includegraphics[width=\linewidth]{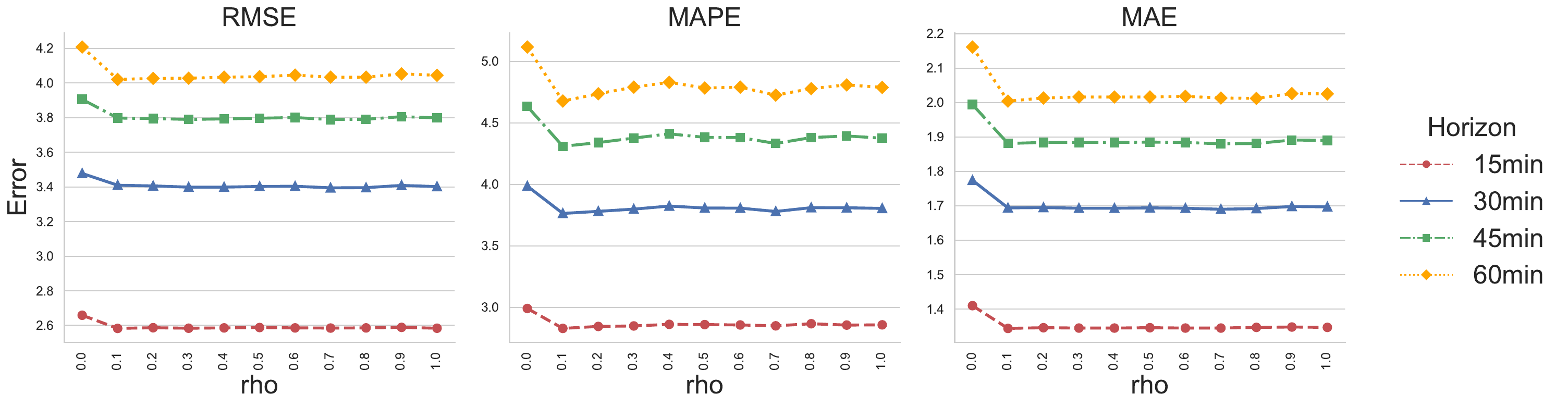}
    \caption{Ablation on different $\rho$ values}
    \label{fig:abl_rho}
\end{figure}

{
The ablation study presented in Figure \ref{fig:abl_rho} examines the impact of different weighting parameters $\rho$ in the loss function equation, using the GWN as the baseline model with kernel size K=3 on the PEMS-BAY 2017 traffic speed dataset. The figure comprehensively shows how three different error metrics vary across the range of $\rho$ values from 0 to 1. When $\rho=0$, the model relies solely on MSE loss, while $\rho=1$ represents exclusive use of NLL loss. The intermediate values represent different weightings in the multi-objective loss function.

First, the results demonstrate a consistent pattern across all three error metrics, where incorporating any amount of NLL loss ($\rho>0$) leads to improved model performance compared to using MSE loss alone. Furthermore, the results also reveal that while including NLL loss is beneficial, setting $\rho<1$ (maintaining some MSE component) yields slightly better results than using pure NLL loss. This improvement is particularly observable for MAPE, which is especially significant as it indicates enhanced prediction accuracy during congested traffic conditions. This suggests that the optimal approach involves a balanced combination of both loss functions, allowing the model to capture both the overall prediction accuracy through MSE and the uncertainty estimation through NLL.
}

\section{Conclusion}
This study proposes a scalable and efficient way to learn the time-varying and multimodal distribution for traffic forecasting errors. The proposed method characterizes the distribution of error as a dynamic mixture of zero-mean Gaussian distributions with full covariance matrices in the component distribution. The $NQ\times NQ$ spatiotemporal covariance matrix for each component is decomposed as a Kronecker product of an $N\times N$ spatial covariance matrix and a $Q\times Q$ temporal covariance matrix. The proposed method can be seamlessly integrated into existing deep-learning models for traffic forecasting as an add-on module. The proposed method is tested using two widely-used traffic forecasting models, GWN and STGCN. The results show that the proposed method can improve forecasting accuracy. Also, the proposed method offers an interpretable framework to understand the traffic pattern by analyzing the learned parameters in the dynamic mixture model.

The proposed method can be extended to other applications where there is a complex high-dimensional correlation in a time-varying manner. In particular, it can be useful to model the prediction error for wind speed forecasting and climate forecasting, since there exists an even larger multidimensional covariance matrix. The scalable solution proposed in this study is expected to be effective in addressing this challenge.

In the course of our study, we encountered several limitations within the proposed method, highlighting key future research directions to focus on and improve.
One of the limitations is the reliance on a multivariate Gaussian distribution as the component distributions in the mixture model. While this offers mathematical convenience and computational efficiency, it may not fully encapsulate the complexities and non-linearities inherent in real-world traffic data. Future research could explore the integration of non-Gaussian distributions or more complex probabilistic models that can better capture the non-linear and non-normal characteristics of traffic flow. 
Another limitations is related to the assumption of a Kronecker product structure for the covariance matrix may not always accurately capture the complex interdependencies in real-world traffic data. This assumption can be a strong assumption since the covariance matrix can have a different structure. As a result, exploring more efficient and effective ways to parameterize the covariance matrix would be a potential research direction. This can lead to more accurate modeling of the complex spatiotemporal dependencies in traffic data. Another research direction is related to the structure of the precision matrix. It could be proposed to use Gaussian Markov Random Fields (GMRF) \cite{rue2005gaussian} by incorporating the road network structure and first-order temporal relationship. This approach would allow for more accurate modeling of the dependencies between different roads in the network. Additionally, an underlying spatiotemporal graph could also be assumed to directly model the sparse spatiotemporal precision ($\Lambda^k \in \mathbb{R}^{NQ\times NQ}$) without assuming a matrix normal distribution. This approach could provide further insights into the relationships between traffic speeds at different locations and at different times.

{
Finally, an important future research direction is to establish a formal \textbf{convergence guarantee} for the proposed method. While the current framework effectively learns the covariance matrix and mixture components in a data-driven manner, similar to previous works on covariance estimation \citep{shukla2024tic, seitzer2022pitfalls, immer2024effective} this paper does not provide a theoretical analysis of its convergence properties. Providing a formal convergence guarantee would offer greater confidence in the stability and robustness of the method, especially in critical applications like traffic forecasting and risk-sensitive domains such as financial markets and energy systems.
}

\section*{Acknowledgement}
The authors acknowledge the financial support of IVADO through its Fundamental Research Funding Program. (Project Title: Bridging Data-Driven and Behavioural Models for Transportation)

\newpage
\bibliographystyle{unsrtnat}
\bibliography{references}

\begin{thebibliography}{39}
\providecommand{\natexlab}[1]{#1}
\providecommand{\url}[1]{\texttt{#1}}
\expandafter\ifx\csname urlstyle\endcsname\relax
  \providecommand{\doi}[1]{doi: #1}\else
  \providecommand{\doi}{doi: \begingroup \urlstyle{rm}\Url}\fi

\bibitem[Li et~al.(2017)Li, Yu, Shahabi, and Liu]{li2017diffusion}
Yaguang Li, Rose Yu, Cyrus Shahabi, and Yan Liu.
\newblock Diffusion convolutional recurrent neural network: Data-driven traffic forecasting.
\newblock \emph{arXiv preprint arXiv:1707.01926}, 2017.

\bibitem[Yu et~al.(2017)Yu, Yin, and Zhu]{yu2017spatio}
Bing Yu, Haoteng Yin, and Zhanxing Zhu.
\newblock Spatio-temporal graph convolutional networks: A deep learning framework for traffic forecasting.
\newblock \emph{arXiv preprint arXiv:1709.04875}, 2017.

\bibitem[Wu et~al.(2019)Wu, Pan, Long, Jiang, and Zhang]{wu2019graph}
Zonghan Wu, Shirui Pan, Guodong Long, Jing Jiang, and Chengqi Zhang.
\newblock Graph wavenet for deep spatial-temporal graph modeling.
\newblock \emph{arXiv preprint arXiv:1906.00121}, 2019.

\bibitem[Cai et~al.(2020{\natexlab{a}})Cai, Janowicz, Mai, Yan, and Zhu]{cai2020traffic}
Ling Cai, Krzysztof Janowicz, Gengchen Mai, Bo~Yan, and Rui Zhu.
\newblock Traffic transformer: Capturing the continuity and periodicity of time series for traffic forecasting.
\newblock \emph{Transactions in GIS}, 24\penalty0 (3):\penalty0 736--755, 2020{\natexlab{a}}.

\bibitem[Li et~al.(2015)Li, Su, Zhang, Lin, and Li]{li2015trend}
Li~Li, Xiaonan Su, Yi~Zhang, Yuetong Lin, and Zhiheng Li.
\newblock Trend modeling for traffic time series analysis: An integrated study.
\newblock \emph{IEEE Transactions on Intelligent Transportation Systems}, 16\penalty0 (6):\penalty0 3430--3439, 2015.

\bibitem[Wilson and Ghahramani(2010)]{wilson2010generalised}
Andrew~Gordon Wilson and Zoubin Ghahramani.
\newblock Generalised wishart processes.
\newblock \emph{arXiv preprint arXiv:1101.0240}, 2010.

\bibitem[Bauwens et~al.(2006)Bauwens, Laurent, and Rombouts]{bauwens2006multivariate}
Luc Bauwens, S{\'e}bastien Laurent, and Jeroen~VK Rombouts.
\newblock Multivariate garch models: a survey.
\newblock \emph{Journal of applied econometrics}, 21\penalty0 (1):\penalty0 79--109, 2006.

\bibitem[Salinas et~al.(2020)Salinas, Flunkert, Gasthaus, and Januschowski]{salinas2020deepar}
David Salinas, Valentin Flunkert, Jan Gasthaus, and Tim Januschowski.
\newblock Deepar: Probabilistic forecasting with autoregressive recurrent networks.
\newblock \emph{International Journal of Forecasting}, 36\penalty0 (3):\penalty0 1181--1191, 2020.

\bibitem[Salinas et~al.(2019)Salinas, Bohlke-Schneider, Callot, Medico, and Gasthaus]{salinas2019high}
David Salinas, Michael Bohlke-Schneider, Laurent Callot, Roberto Medico, and Jan Gasthaus.
\newblock High-dimensional multivariate forecasting with low-rank gaussian copula processes.
\newblock \emph{Advances in neural information processing systems}, 32, 2019.

\bibitem[Vaughan(1970)]{vaughan1970distribution}
Rodney~J Vaughan.
\newblock The distribution of traffic volumes.
\newblock \emph{Transportation Science}, 4\penalty0 (1):\penalty0 97--110, 1970.

\bibitem[Kerner and Rehborn(1997)]{kerner1997experimental}
Boris~S Kerner and Hubert Rehborn.
\newblock Experimental properties of phase transitions in traffic flow.
\newblock \emph{Physical Review Letters}, 79\penalty0 (20):\penalty0 4030, 1997.

\bibitem[Couton et~al.(1997)Couton, Danech-Pajouh, et~al.]{couton1997application}
Fran{\c{c}}ois Couton, Medhi Danech-Pajouh, et~al.
\newblock Application of the mixture of probability distributions to the recognition of road traffic flow regimes.
\newblock \emph{IFAC Proceedings Volumes}, 30\penalty0 (8):\penalty0 715--720, 1997.

\bibitem[Park et~al.(2010)Park, Zhang, and Lord]{park2010bayesian}
Byung-Jung Park, Yunlong Zhang, and Dominique Lord.
\newblock Bayesian mixture modeling approach to account for heterogeneity in speed data.
\newblock \emph{Transportation research part B: methodological}, 44\penalty0 (5):\penalty0 662--673, 2010.

\bibitem[Wong and Li(2000)]{wong2000mixture}
Chun~Shan Wong and Wai~Keung Li.
\newblock On a mixture autoregressive model.
\newblock \emph{Journal of the Royal Statistical Society: Series B (Statistical Methodology)}, 62\penalty0 (1):\penalty0 95--115, 2000.

\bibitem[Bishop(1994)]{bishop1994mixture}
Christopher~M Bishop.
\newblock Mixture density networks.
\newblock 1994.

\bibitem[Nikolaev et~al.(2013)Nikolaev, Tino, and Smirnov]{nikolaev2013time}
Nikolay Nikolaev, Peter Tino, and Evgueni Smirnov.
\newblock Time-dependent series variance learning with recurrent mixture density networks.
\newblock \emph{Neurocomputing}, 122:\penalty0 501--512, 2013.

\bibitem[Ellefsen et~al.(2019)Ellefsen, Martin, and Torresen]{ellefsen2019mixture}
Kai~Olav Ellefsen, Charles~Patrick Martin, and Jim Torresen.
\newblock How do mixture density rnns predict the future?
\newblock \emph{arXiv preprint arXiv:1901.07859}, 2019.

\bibitem[Mao et~al.(2023)Mao, Wan, Wen, Wu, Zheng, Qiang, Guo, Wu, Hu, and Lin]{mao2023gmdnet}
Xiaowei Mao, Huaiyu Wan, Haomin Wen, Fan Wu, Jianbin Zheng, Yuting Qiang, Shengnan Guo, Lixia Wu, Haoyuan Hu, and Youfang Lin.
\newblock Gmdnet: A graph-based mixture density network for estimating packages’ multimodal travel time distribution.
\newblock In \emph{Proceedings of the AAAI Conference on Artificial Intelligence}, volume~37, pages 4561--4568, 2023.

\bibitem[Chen et~al.(2021)Chen, Chen, Cai, Li, Guo, and Li]{chen2021short}
Mingjian Chen, Rui Chen, Fu~Cai, Wanli Li, Naikun Guo, and Guangyun Li.
\newblock Short-term traffic flow prediction with recurrent mixture density network.
\newblock \emph{Mathematical Problems in Engineering}, 2021:\penalty0 1--9, 2021.

\bibitem[Li et~al.(2023)Li, Normandin-Taillon, Wang, and Huang]{li2023xrmdn}
Xiaoming Li, Hubert Normandin-Taillon, Chun Wang, and Xiao Huang.
\newblock Xrmdn: A recurrent mixture density networks-based architecture for short-term probabilistic demand forecasting in mobility-on-demand systems with high volatility.
\newblock \emph{arXiv preprint arXiv:2310.09847}, 2023.

\bibitem[Liu et~al.(2023)Liu, Lyu, Wang, Wang, Liu, and Meng]{liu2023gaussian}
Zhiyuan Liu, Cheng Lyu, Zelin Wang, Shuaian Wang, Pan Liu, and Qiang Meng.
\newblock A gaussian-process-based data-driven traffic flow model and its application in road capacity analysis.
\newblock \emph{IEEE Transactions on Intelligent Transportation Systems}, 24\penalty0 (2):\penalty0 1544--1563, 2023.

\bibitem[Cai et~al.(2020{\natexlab{b}})Cai, Cheng, Ding, Chen, Li, and Vucetic]{cai2020spatiotemporal}
Yue Cai, Peng Cheng, Ming Ding, Youjia Chen, Yonghui Li, and Branka Vucetic.
\newblock Spatiotemporal gaussian process kalman filter for mobile traffic prediction.
\newblock In \emph{2020 IEEE 31st Annual International Symposium on Personal, Indoor and Mobile Radio Communications}, pages 1--6. IEEE, 2020{\natexlab{b}}.

\bibitem[Bayati et~al.(2020)Bayati, Nguyen, and Cheriet]{bayati2020gaussian}
Abdolkhalegh Bayati, Kim-Khoa Nguyen, and Mohamed Cheriet.
\newblock Gaussian process regression ensemble model for network traffic prediction.
\newblock \emph{IEEE Access}, 8:\penalty0 176540--176554, 2020.

\bibitem[Hara et~al.(2018)Hara, Suzuki, and Kuwahara]{hara2018network}
Yusuke Hara, Junpei Suzuki, and Masao Kuwahara.
\newblock Network-wide traffic state estimation using a mixture gaussian graphical model and graphical lasso.
\newblock \emph{Transportation Research Part C: Emerging Technologies}, 86:\penalty0 622--638, 2018.

\bibitem[Lei et~al.(2022)Lei, Labbe, Wu, and Sun]{lei2022bayesian}
Mengying Lei, Aurelie Labbe, Yuankai Wu, and Lijun Sun.
\newblock Bayesian kernelized matrix factorization for spatiotemporal traffic data imputation and kriging.
\newblock \emph{IEEE Transactions on Intelligent Transportation Systems}, 23\penalty0 (10):\penalty0 18962--18974, 2022.

\bibitem[Greenewald and Hero(2015)]{greenewald2015robust}
Kristjan Greenewald and Alfred~O Hero.
\newblock Robust kronecker product pca for spatio-temporal covariance estimation.
\newblock \emph{IEEE Transactions on Signal Processing}, 63\penalty0 (23):\penalty0 6368--6378, 2015.

\bibitem[Kariya and Kurata(2004)]{kariya2004generalized}
Takeaki Kariya and Hiroshi Kurata.
\newblock \emph{Generalized least squares}.
\newblock John Wiley \& Sons, 2004.

\bibitem[Fausett and Fulton(1994)]{fausett1994large}
Donald~W Fausett and Charles~T Fulton.
\newblock Large least squares problems involving kronecker products.
\newblock \emph{SIAM Journal on Matrix Analysis and Applications}, 15\penalty0 (1):\penalty0 219--227, 1994.

\bibitem[Karimi and Dehghan(2018)]{karimi2018global}
Saeed Karimi and Maryam Dehghan.
\newblock Global least squares method based on tensor form to solve linear systems in kronecker format.
\newblock \emph{Transactions of the Institute of Measurement and Control}, 40\penalty0 (7):\penalty0 2378--2386, 2018.

\bibitem[Marco et~al.(2019)Marco, Mart{\'\i}nez, and Via{\~n}a]{marco2019least}
Ana Marco, Jos{\'e}-Javier Mart{\'\i}nez, and Raquel Via{\~n}a.
\newblock Least squares problems involving generalized kronecker products and application to bivariate polynomial regression.
\newblock \emph{Numerical Algorithms}, 82:\penalty0 21--39, 2019.

\bibitem[Yang et~al.(2019)Yang, Fan, and Royset]{yang2019estimating}
Y.~Yang, Y.~Fan, and J.~O. Royset.
\newblock Estimating probability distribution of travel demand on congested networks.
\newblock \emph{Transportation Research Part B: Methodological}, 122:\penalty0 265--286, 2019.

\bibitem[Chen et~al.(2001)Chen, Petty, Skabardonis, Varaiya, and Jia]{chen2001freeway}
Chao Chen, Karl Petty, Alexander Skabardonis, Pravin Varaiya, and Zhanfeng Jia.
\newblock Freeway performance measurement system: mining loop detector data.
\newblock \emph{Transportation Research Record}, 1748\penalty0 (1):\penalty0 96--102, 2001.

\bibitem[Jagadish et~al.(2014)Jagadish, Gehrke, Labrinidis, Papakonstantinou, Patel, Ramakrishnan, and Shahabi]{jagadish2014big}
Hosagrahar~V Jagadish, Johannes Gehrke, Alexandros Labrinidis, Yannis Papakonstantinou, Jignesh~M Patel, Raghu Ramakrishnan, and Cyrus Shahabi.
\newblock Big data and its technical challenges.
\newblock \emph{Communications of the ACM}, 57\penalty0 (7):\penalty0 86--94, 2014.

\bibitem[Li et~al.(2022)Li, Knoop, and van Lint]{li2022estimate}
Guopeng Li, Victor~L Knoop, and Hans van Lint.
\newblock Estimate the limit of predictability in short-term traffic forecasting: An entropy-based approach.
\newblock \emph{Transportation Research Part C: Emerging Technologies}, 138:\penalty0 103607, 2022.

\bibitem[Chen et~al.(2023)Chen, Cheng, Jin, Tr{\'e}panier, and Sun]{chen2023probabilistic}
Xiaoxu Chen, Zhanhong Cheng, Jian~Gang Jin, Martin Tr{\'e}panier, and Lijun Sun.
\newblock Probabilistic forecasting of bus travel time with a bayesian gaussian mixture model.
\newblock \emph{Transportation Science}, 57\penalty0 (6):\penalty0 1516--1535, 2023.

\bibitem[Rue and Held(2005)]{rue2005gaussian}
Havard Rue and Leonhard Held.
\newblock \emph{Gaussian Markov random fields: theory and applications}.
\newblock Chapman and Hall/CRC, 2005.

\bibitem[Shukla et~al.(2024)Shukla, Salzmann, and Alahi]{shukla2024tic}
Megh Shukla, Mathieu Salzmann, and Alexandre Alahi.
\newblock Tic-tac: A framework for improved covariance estimation in deep heteroscedastic regression.
\newblock In \emph{Proceedings of the 41st International Conference on Machine Learning (ICML) 2024}, 2024.

\bibitem[Seitzer et~al.(2022)Seitzer, Tavakoli, Antic, and Martius]{seitzer2022pitfalls}
Maximilian Seitzer, Arash Tavakoli, Dimitrije Antic, and Georg Martius.
\newblock On the pitfalls of heteroscedastic uncertainty estimation with probabilistic neural networks.
\newblock \emph{arXiv preprint arXiv:2203.09168}, 2022.

\bibitem[Immer et~al.(2024)Immer, Palumbo, Marx, and Vogt]{immer2024effective}
Alexander Immer, Emanuele Palumbo, Alexander Marx, and Julia Vogt.
\newblock Effective bayesian heteroscedastic regression with deep neural networks.
\newblock \emph{Advances in Neural Information Processing Systems}, 36, 2024.

\end{thebibliography}

\end{document}